\documentclass{article} % For LaTeX2e
\usepackage{iclr2019_conference,times}

\usepackage{amsmath,amsfonts,bm}

\def\secref#1{section~\ref{#1}}
\def\eqref#1{equation~\ref{#1}}
\def\1{\bm{1}}

\DeclareMathAlphabet{\mathsfit}{\encodingdefault}{\sfdefault}{m}{sl}
\SetMathAlphabet{\mathsfit}{bold}{\encodingdefault}{\sfdefault}{bx}{n}

\DeclareMathOperator*{\argmax}{arg\,max}

\usepackage{hyperref}
\usepackage{url}
\usepackage{graphicx, subcaption}
\usepackage{todonotes, soul}

\usepackage{booktabs} % Nice tables

\newcommand*{\affmark}[1][*]{\textsuperscript{#1}}
\newcommand*{\affaddr}[1]{#1}

\title{GAN-based Generation and Automatic Selection of Explanations for Neural Networks}

\author{Saumitra Mishra\affmark[*], Daniel Stoller\affmark[*], Emmanouil Benetos\affmark[* \#], Bob L. Sturm\affmark[\dag] \& Simon Dixon\affmark[*] 
\\\\
\affaddr{\affmark[*]School of EECS, Queen Mary University of London, UK}\\
\affaddr{\affmark[\#]The Alan Turing Institute, UK}\\
\affaddr{\affmark[\dag]Speech, Music and Hearing, KTH Royal Institute of Technology, Sweden} \\
\texttt{\{saumitra.mishra, d.stoller, emmanouil.benetos\}@qmul.ac.uk}, \\
\texttt{bobs@kth.se, s.e.dixon@qmul.ac.uk}
}

\iclrfinalcopy % Uncomment for camera-ready version, but NOT for submission.
\begin{document}

\maketitle

\begin{abstract}
One way to interpret trained deep neural networks (DNNs) is by inspecting characteristics that neurons in the model respond to, such as by iteratively optimising the model input (e.g., an image) to maximally activate specific neurons.
However, this requires a careful selection of hyper-parameters to generate interpretable examples for each neuron of interest, and current methods rely on a manual, qualitative evaluation of each setting, which is prohibitively slow.
We introduce a new metric that uses Fr\'{e}chet Inception Distance (FID) to encourage similarity between model activations for real and generated data.
This provides an efficient way to evaluate a set of generated examples for each setting of hyper-parameters.
We also propose a novel GAN-based method for generating explanations that enables an efficient search through the input space and imposes a strong prior favouring realistic outputs.
We apply our approach to a classification model trained to predict whether a music audio recording contains singing voice. Our results suggest that this proposed metric successfully selects hyper-parameters leading to interpretable examples, avoiding the need for manual evaluation.
Moreover, we see that examples synthesised to maximise or minimise the predicted probability of singing voice presence exhibit vocal or non-vocal characteristics, respectively, suggesting that our approach is able to generate suitable explanations for understanding concepts learned by a neural network.
\end{abstract}

\section{Introduction}

There is an increasing interest in interpreting black-box machine learning models, especially Deep Neural Networks (DNNs)~\citep{DoshiKim2017}.
Insights about how models function can assist in gaining trust in their predictions -- an essential factor for model adoption in safety-critical applications (e.g. health care, self-driving cars)~\citep{Ribeiro_kdd_2016}.
We can understand a machine learning model by employing one of two strategies. The first involves training inherently interpretable models and is a promising research direction, but often such models perform poorly when compared to state-of-the-art black-box models~\citep{Ribeiro_icml_2016}.
In contrast, the second strategy involves a post-hoc analysis of a trained model, which does not require compromises on its predictive capacity.

There are two key approaches to bring post-hoc interpretability to DNNs in particular~\citep{Montavon2018}. 
The first focuses on explaining the predictions of a model \citep{Simonyan2014, Bach2015, Ribeiro_kdd_2016} for a given input, while the second analyses components (e.g. neurons or layers) of a DNN~\citep{Olah2017}.
We focus on the second approach in this paper, as it yields general insights about how a DNN forms its predictions.

We can analyse components of a DNN using different methodologies. For example, one can use feature inversion to map latent codes to the input space highlighting the discriminative information a DNN preserves at its layers \citep{Mahendran2015, Dosovitskiy2016_cvpr}. In another direction, one can analyse features that different components of a DNN are sensitive to. One way to do this is by identifying instances from the dataset that maximally activate different components in a DNN \citep{Zhou2015_iclr}. 
Another way is by using Activation Maximisation (AM)~\citep{Erhan2009} that iteratively optimises random noise to synthesise examples in the input space (e.g., images) to maximally activate a neuron or layer in a DNN.
Since AM is data-independent and tends to focus more on the explanatory input factors, we will pursue an AM-based approach in this paper.

The interpretability of examples generated by AM depends on two key factors: optimisation of hyper-parameters and the prior. Generally, interpretable examples are selected for each neuron by performing a grid search in the hyper-parameter space and visually inspecting each generated example~\citep{Nguyen_nips_2016}, but this is subjective, prohibitively slow and limits the hyper-parameter search space. Also, such an approach is not scalable to analysing other DNN neurons that may require different hyper-parameter settings.  

The use of priors for AM restricts the input search space to prevent generating uninformative, adversarial examples.
Researchers have proposed several hand-crafted priors for effective AM, synthesising interpretable images \citep{Yosinski2015, Nguyen_iclr_2016, Mahendran2015}.
In another direction, \citet{Nguyen_nips_2016} demonstrated that replacing hand-crafted priors by a learned prior (adversarially trained feature inverter) considerably improves the interpretability of synthesised images.
However, their approach requires training a separate prior for each layer in the classifier, and appears to rely on the prior and the classifier model having similar architectures.

\begin{figure}[t]
\begin{center}
\framebox{\includegraphics[width=0.7\textwidth]{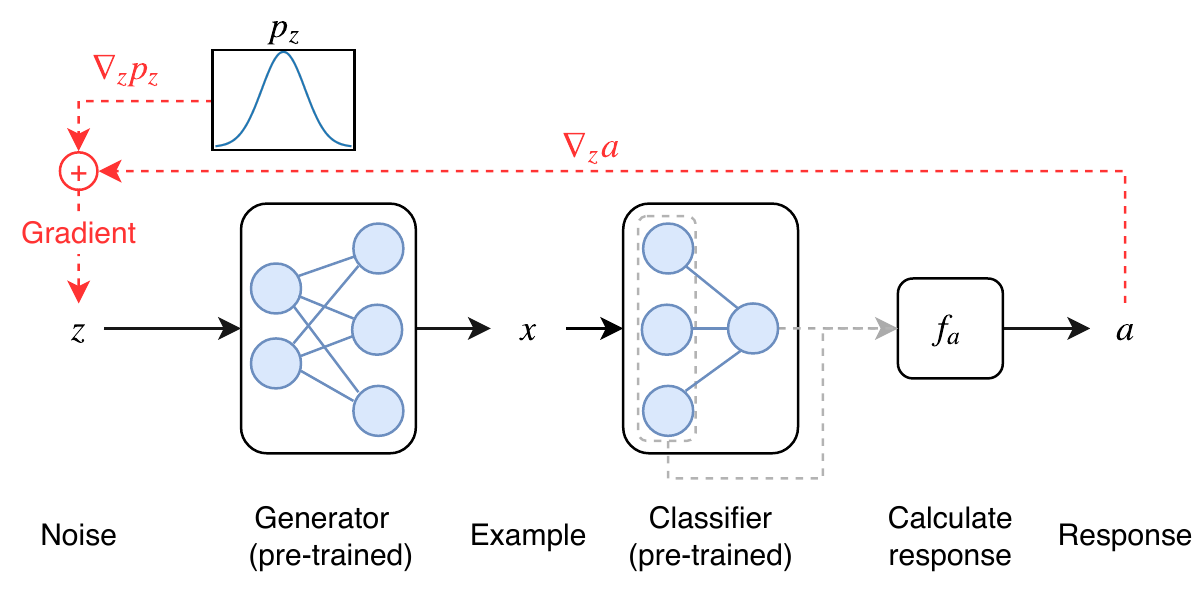}}
\end{center}
\caption{Overview of our proposed approach. A noise vector $z$ is used to generate an example $x$, for which a response $a \in \mathbb{R}$ is calculated with a response function $f_a$ from all neuron activations of the classifier. $f_a$ can be defined depending on which aspect of the classifier is of interest; examples include the activation of a certain neuron, or the average layer activation.
$z$ is optimised to maximise the response $a$, but also the prior probability $p_z(z)$ to favour realistic outputs.}
\label{fig:overview}
\end{figure}

In this work, we aim to tackle these challenges, making the following contributions:
\begin{itemize}
\item To our knowledge, we are the first to use a Generative Adversarial Network (GAN) for example generation using AM. This imposes a strong prior and enables effective AM for any given part of a classifier and even other classifiers with the same input domain without re-training the generator. The work by \citet{Nguyen2017_cvpr} is closest to ours, in which the authors use a denoising autoencoder as a prior on the latent code of the adversarially trained feature inverter.
\item We propose a quantitative measure estimating the interpretability of a set of generated examples by adopting the Fr\'{e}chet Inception Distance (FID) metric \citep{Heusel2017}. We provide evidence for its effectiveness by qualitatively analysing the synthesised examples. 
\item We apply our method to a state-of-the-art deep audio classification model that predicts singing voice activity in music excerpts. This results in visualisations that successfully capture the concept represented by the ground truth labels the classifier was trained to predict. There have been some recent works in understanding deep audio classification models, but they either use a different method \citep{Mishra_ismir_2018} or perform AM with hand-crafted priors \citep{Zhang_icassp_2018, Krun2018}. 

\end{itemize}

\section{Method}

Figure~\ref{fig:overview} provides an overview of our method.
For a pre-trained neural network classifier $f_c$ with $M$ neurons and input $x \in \mathbb{R}^d$, our goal is to provide examples that activate a given neuron activation pattern (``classifier response'').
Formally, we define $f_n(x) \in \mathbb{R}^M$ as the output activations of \emph{all} $M$ neurons in the classifier $f_c$ for a given input example $x$.
The classifier response we aim to explain can then be defined in a general fashion as the output of some function $f_a : \mathbb{R}^M \rightarrow \mathbb{R}$ that takes \emph{all} $M$ neuron activations of the classifier as input.
$f_a$ can be set to output the activation of a single neuron, or the average activation of one or multiple layers, but any differentiable function is supported.

\subsection{Activation maximisation}

We can perform activation maximisation to find an input example $\hat{x} \in \mathbb{R}^d$ so that the resulting activation $f_a(\cdot)$ is maximised:
\begin{equation}
    \hat{x} = \argmax_x f_a(f_n(x))
\label{eq:am}
\end{equation}
The above objective can be optimised by stochastic gradient descent (SGD) by backpropagating through the classifier layers.

\subsection{GAN-based prior}
\label{sec:gan_prior}

However, activation maximisation often produces adversarial examples~\citep{Nguyen2015}, which can be very different from inputs encountered during classifier training and testing, are hard to interpret and do not explain the classifier's behaviour for real-world inputs.
Furthermore, optimising over the input $x$ directly is often difficult, especially if the dimensionality $d$ is high~\citep{Nguyen_nips_2016}.

Our method makes use of a GAN (for more details, see~\citet{Goodfellow2014}), where a generator $f_g : \mathbb{R}^n \rightarrow \mathbb{R}^d$ is trained to map a noise vector $z \in \mathbb{R}^n$ drawn from a known noise distribution $p_z$ to a generated example $x$, and optimises
\begin{equation}
    \hat{z} = \argmax_z f_a(f_n(f_g(z))) + \lambda \log p_z(z).
\label{eq:am_gan}
\end{equation}
The weighting term $\lambda \geq 0$ is a hyper-parameter controlling the trade-off between activation maximisation and the realism of the generated examples.
Note that we search in the low-dimensional noise space for a vector $\hat{z}$ whose associated generator output $f_g(\hat{z})$ produces a high activation, which avoids optimisation issues.
To encourage realistic outputs, the real data density $p_x$ should ideally be used in the form of a prior term $\log p_x(f_g(z))$ in~\eqref{eq:am_gan}, but we do not have access to $p_x$.
However, assuming a well-trained generator, we can use $\log p_z(z)$ instead, since it should be approximately proportional.

To optimise~\eqref{eq:am_gan} with gradient descent, we require $p_z$ to be a continuously differentiable distribution.
Note that this does not include the uniform distribution commonly used for training GANs, for example in \citep{Goodfellow2014,Radford2015,Hjelm2017}.

\subsection{Example generation}

The previous~\secref{sec:gan_prior} demonstrated how one example is generated in our approach.
To generate a set of $N$ examples, we draw $N$ random noise vectors $\tilde{z}_1,\ldots,\tilde{z}_N$ independently from $p_z$ as initialisation points for SGD.
The resulting examples should be diverse, so converging to the same optima of~\eqref{eq:am_gan} independent of initialisation is undesirable.
Therefore we set the SGD learning rate $l_r$ as well as the number of update steps $N_t$ as hyper-parameters, since they control the influence of the initialisation points on the generated examples and thereby the amount of diversity and randomness.

\subsection{Hyper-parameter optimisation}
\label{sec:hyperparam}

\begin{figure}[t]
\begin{center}
\includegraphics[height=0.25\textwidth]{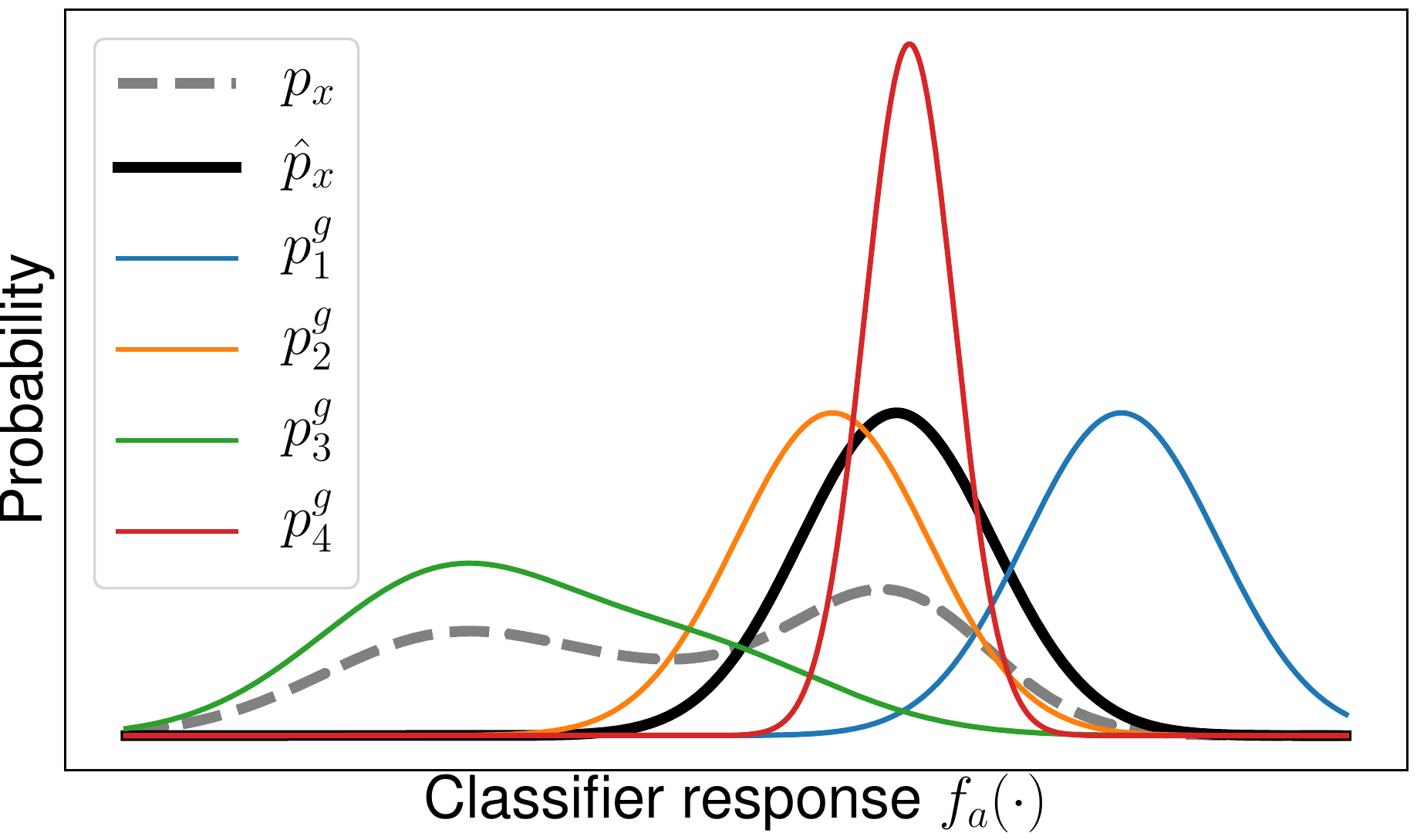}
\end{center}
\caption{Intuitive explanation for our proposed metric, showing the distributions of activations $f_a(\cdot)$ obtained for input examples from the dataset ($p_x$), of the dataset examples with the highest $N$ responses $f_a(\cdot)$ ($\hat{p}_x$), and of four hypothetical generators, $p^g_1, \ldots. p^g_4$. Our metric determines which generator distribution is most similar to $\hat{p}_x$ to ensure realistic examples.
}
\label{fig:venn}
\end{figure}

To optimise the prior weight $\lambda$ and optimisation parameters $l_r$ as well as $N_t$, it would be ideal to have human subjects evaluate the usefulness of the explanations resulting from different configurations, but this is prohibitively time-intensive.
Therefore, we introduce a novel, automatic metric for quickly evaluating a set of generated explanations, allowing efficient hyper-parameter optimisation.
In the following, we will explain our reasoning using the hypothetical example in Figure~\ref{fig:venn}.
We posit that good interpretability requires the generated examples to have a similar distribution of classifier responses $f_a(\cdot)$ as the $N$ samples with the highest response from the dataset ($\hat{p}_x$ in Figure~\ref{fig:venn}).
This is because unrealistic adversarial examples (generator $1$ in Figure~\ref{fig:venn}) often lead to large responses.
Also, too much weight on the GAN prior $\lambda$ or ineffective optimisation (generator $3$) leads to examples that are realistic, but have too low responses compared to real examples.
Additionally, the variance of responses should be similar (making generator $2$ the best according to our metric) to ensure a sufficient degree of diversity in the generated samples (in contrast to generator $4$).

To take the average and the variance of the responses into account, we therefore adopt the Fr\'{e}chet Inception Distance (FID)~\citep{Heusel2017} as our distance metric.
Since our responses $f_a$ for each example are scalar values, the FID reduces to
\begin{equation}
    D((\mu_r,\sigma_r),(\mu_g,\sigma_g)) = (\mu_r - \mu_g)^2 + \sigma_r + \sigma_g - 2  (\sigma_r  \sigma_g)^{\frac{1}{2}},
\end{equation}
where $\mu_r$ and $\mu_g$ are the means and $\sigma_r$ and $\sigma_g$ the unbiased sample variance of the (one-dimensional) real and the generated response distribution, respectively.

In the following~\secref{sec:experiments}, we will investigate whether the metric proposed above adequately reflects how useful the set of generated examples is to a human observer.

\section{Experiments}
\label{sec:experiments}

To evaluate our method and to investigate whether approaches based on AM can transfer to domains other than computer vision, we apply it to audio classification.
Specifically, we will consider Singing Voice Detection (SVD), a binary classification task~\citep{Lee2018} where a classifier predicts whether singing voice (vocals) is present in a segment of a music recording.

\subsection{Choice of classifier}
\label{sec:classifier}

We select a state-of-the-art SVD model\footnote{Available as open source at \url{https://github.com/f0k/ismir2015}} introduced by~\citet{Schluter2015}.
The model is an eight-layer Convolutional Neural Network (CNN) the architecture of which is mentioned in Appendix (Table~\ref{table:svdnet_arch}). 
It takes a Mel spectrogram of an audio excerpt of $1.6$s duration as input. 
The Mel spectrogram is calculated by first applying an FFT~\footnote{Using a window size of $1024$ and a hop size of $315$ samples, with audio sampled at $22050$Hz} and taking the magnitudes of the resulting spectrum. Later, we apply a Mel filterbank~\footnote{The Mel filterbank is a set of band-pass filters distributed along the Mel-frequency scale, which is a perceptual scale of pitch defining a logarithmic relationship between frequency and perceived pitch. We use $80$ filters in our work, ranging from $27.5$ to $8000$ Hz.} to summarise the energies across different frequency bands.
Finally, we normalise the resulting non-negative values by applying $x \rightarrow \log (\max(x, 10^{-7}))$.
Using a single neuron with sigmoid activation in the last layer, the CNN predicts the probability of singing voice being present at the centre of the input audio excerpt. 
\citet{Schluter2015} train the model in a supervised fashion by minimising the binary cross-entropy loss between model predictions and the ground-truth labels.

As training input, the authors randomly sample excerpts from the 93 French Pop music songs contained in the Jamendo dataset~\citep{Ramona2008}.
The dataset is pre-partitioned into subsets of 61 (training), 16 (validation) and 16 (testing) songs, respectively, and each song has manual annotations indicating the start and end times of each vocal segment.

We replicated the proposed approach of the authors\footnote{The open-sourced version of the classifier introduced by~\citet{Schluter2015} is based on Theano and Lasagne and was ported to Tensorflow as part of this work}, obtaining a classifier whose performance is very similar to the one reported by the authors, as shown in Appendix (Table~\ref{table:svdnet_compare}).

\subsection{Choice of response function}

In this study, we focus on generating positive and negative examples that maximally or minimally excite the final output neuron of the classifier, respectively.
Compared to using other definitions of $f_a$, this allows us to directly evaluate the characteristics of our generated examples, as the positive examples should differ from the negative ones by the presence of singing voice since the classifier is known to be accurate at singing voice detection.

Our initial experiments show that the predicted probability converges to $0$ or $1$ after only very few iterations, leading to vanishing gradients due to saturation of the sigmoid non-linearity, effectively halting optimisation.
We argue this indicates an inherent problem of the classifier and not our method, as neural networks are well-known to be prone to making over-confident predictions~\citep{Gal2016}.
Thus, we applied our method to the pre-sigmoid activations of the final neuron instead.

\subsection{GAN training}

We use the FMA dataset~\citep{Defferrard2017} for training the GAN, selecting only Pop music pieces to reduce the data complexity and to make the song selection more similar to the one used for training the classifier.
The audio signals are converted to Mel spectrograms, replicating the classifier preprocessing described in~\secref{sec:classifier}.
From each song's full spectrogram, we create clips with $115$ time frames and an equal amount of spacing between each clip.

\begin{table}[t]
\footnotesize
\caption{The architecture of our generator. The transposed convolutional layers (ConvT) as well as the FC layer have LeakyReLU activations. 
}
\label{table:gen_arch}
\begin{center}
\begin{tabular}{cccccc} 
 \toprule
  Layer & Input shape & Filter size & Stride & No. of filters/neurons & Output shape \\
 \midrule
 FC & $128$ & - & - & $5120$ & $5120$ \\ 
 ConvT & $8 \times 5 \times 128$ & $5\times 10$ & $2\times 2$ & $64$ & $16 \times 10 \times 64$ \\
 ConvT & $16 \times 10 \times 64$ & $5\times 20$ & $2\times 2$ & $32$ & $32 \times 20 \times 32$ \\
 ConvT & $32 \times 20 \times 32$ & $5\times 20$ & $2\times 2$ & $16$ & $64 \times 40 \times 16$ \\
 ConvT & $64 \times 40 \times 18$ & $5\times 20$ & $2\times 2$ & $8$ & $128 \times 80 \times 8$ \\
 Conv & $128 \times 80 \times 8$ & $5 \times 5$ & $1 \times 1$ & $1$ & $128 \times 80 \times 1$ \\
 \bottomrule
\end{tabular}
\end{center}
\end{table}

\begin{table}[t]
\footnotesize
\caption{The architecture of our discriminator. Convolutional layers have LeakyReLU activations and bias. The fully connected layer has no bias or activation function.
}
\label{table:disc_arch}
\begin{center}
\begin{tabular}{cccccc} 
 \toprule
  Layer & Input shape & Filter size & Stride & No. of filters/neurons & Output shape \\
 \midrule
 Conv & $ 128 \times 80 \times 1 $ & $5\times 80$ & $2\times 2$ & 32 & $64 \times 40 \times 32$ \\
 Conv & $ 64 \times 40 \times 32 $ & $5\times 40$ & $2\times 2$ & 64 & $32 \times 20 \times 64$ \\
 Conv & $ 32 \times 20 \times 64 $ & $5\times 20$ & $2\times 2$ & 128 & $16 \times 10 \times 128$ \\ 
 Conv & $ 16 \times 10 \times 128 $ & $5\times 10$ & $2\times 2$ & 256 & $8 \times 5 \times 256$ \\ 
 FC & $10240$ & - & - &  1 & 1\\ 
 \bottomrule
\end{tabular}
\end{center}
\end{table}

For the generator, we choose a standard normal likelihood $\mathcal{N}(z|\mathbf{0}_n ; \mathbf{I}_n)$ for the continuously differentiable noise term $p_z(z)$, with a dimensionality of $n=128$.
The generator architecture is a CNN adapted from the DCGAN~\citep{Radford2015} and is shown in Table~\ref{table:gen_arch}, using multiple strided transposed convolutions.
The final convolution outputs a $128 \times 80 \times 1$ tensor, which is cropped evenly at the borders to obtain $115$ time frames as required by the classifier.
The final convolution employs $x \rightarrow \max(x, \log(10^{-7}))$ as activation function to ensure the generated spectrogram magnitudes are in the same interval range as the Mel spectrograms obtained from preprocessing real audio samples following~\secref{sec:classifier}.

The discriminator is again similar to the DCGAN~\citep{Radford2015} and is shown in Table~\ref{table:disc_arch}, making use of multiple strided 2D convolutions to process the Mel spectrogram input of size $115 \times 80 \times 1$.
The output is a scalar real value used to distinguish real from generated samples.

%\paragraph{Training procedure}
We use the WGAN-GP objective for training our GAN as in~\citep{Gulrajani2017}, with a GP weight of $10$.
The Adam optimiser with a learning rate of $10^{-4}$ is used to train the generator and discriminator for 600,000 iterations with a batch size of $16$.

\subsection{AM Optimisation}

We perform a grid search over our hyper-parameters, using $l_r \in \{0.1, 0.01, 0.001\}$, $\lambda \in \{0.1, 0.01, 0.001\}$ and $N_t \in \{100, 500, 1000\}$, giving $27$ possible settings.
We sample $N=50$ noise vectors from the noise distribution $p_z$, resulting in $N=50$ examples along with their respective activation values $f_a(\cdot)$ for each setting after applying our method.
Also we feed the training dataset to the classifier and record the last neuron activation for each excerpt, and select the top $N=50$ excerpts with maximum activation. We generate a new excerpt of $115$ consecutive Mel spectrogram frames ($\widehat{=} 1.6$ sec) for every $50$ time frames ($\widehat{=} 0.7$ sec) in a recording.
We optimise our objective in~\eqref{eq:am_gan} by using the Adam optimiser with $\beta_1=0.99$,  $\beta_2=0.999$ and $\epsilon=10^{-8}$.

\section{Results}
\label{sec:results}

In~\secref{sec:results_hyperparam}, we analyse the effectiveness of our metric proposed in~\secref{sec:hyperparam} in the context of hyper-parameter optimisation, before employing the best configuration to evaluate our explanation generation system in~\secref{sec:results_classifier}.

\begin{figure*}[t!]
\begin{center}
{\includegraphics[width=\textwidth]{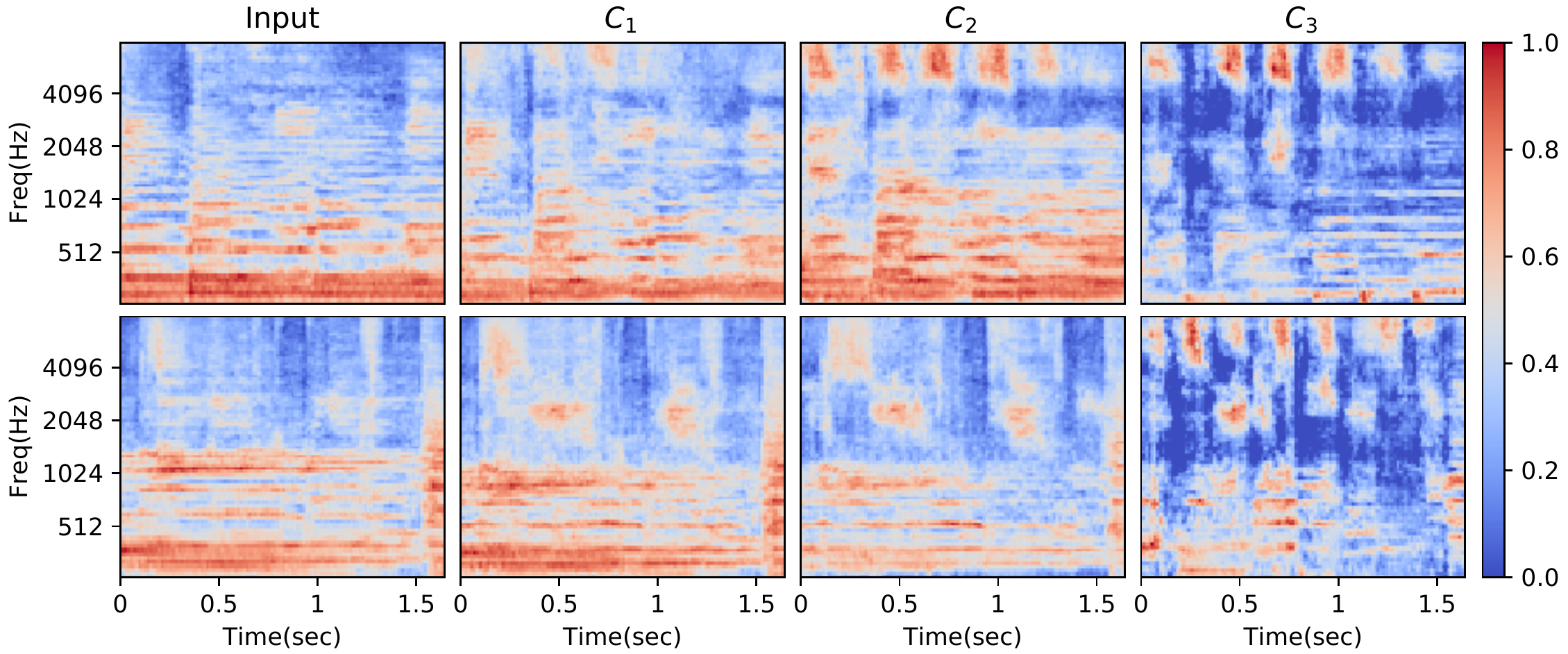}}
\end{center}
 \caption{Mel spectrogram visualisations demonstrating the effectiveness of our example evaluation metric from~\secref{sec:hyperparam}, each normalised in scale independently so that red colors show relatively high and blue colors show relatively low spectral energy. The leftmost column shows the output of our GAN $f_g(\tilde{z}_i)$ for two initial noise vectors $\tilde{z}_1, \tilde{z}_2$ (one per row). The others show the result of applying our method with hyper-parameter configurations $C_1, C_2$ and $C_3$, which represent the best, median and worst configuration from the set of $27$ configurations according to our evaluation metric, respectively. For more details about the configurations, refer to Table~\ref{table:hyperparams} in the Appendix.}
 \label{fig:hyperparam}
 \end{figure*}

 \begin{figure*}[t!]
 \begin{center}
{\includegraphics[width=\textwidth]{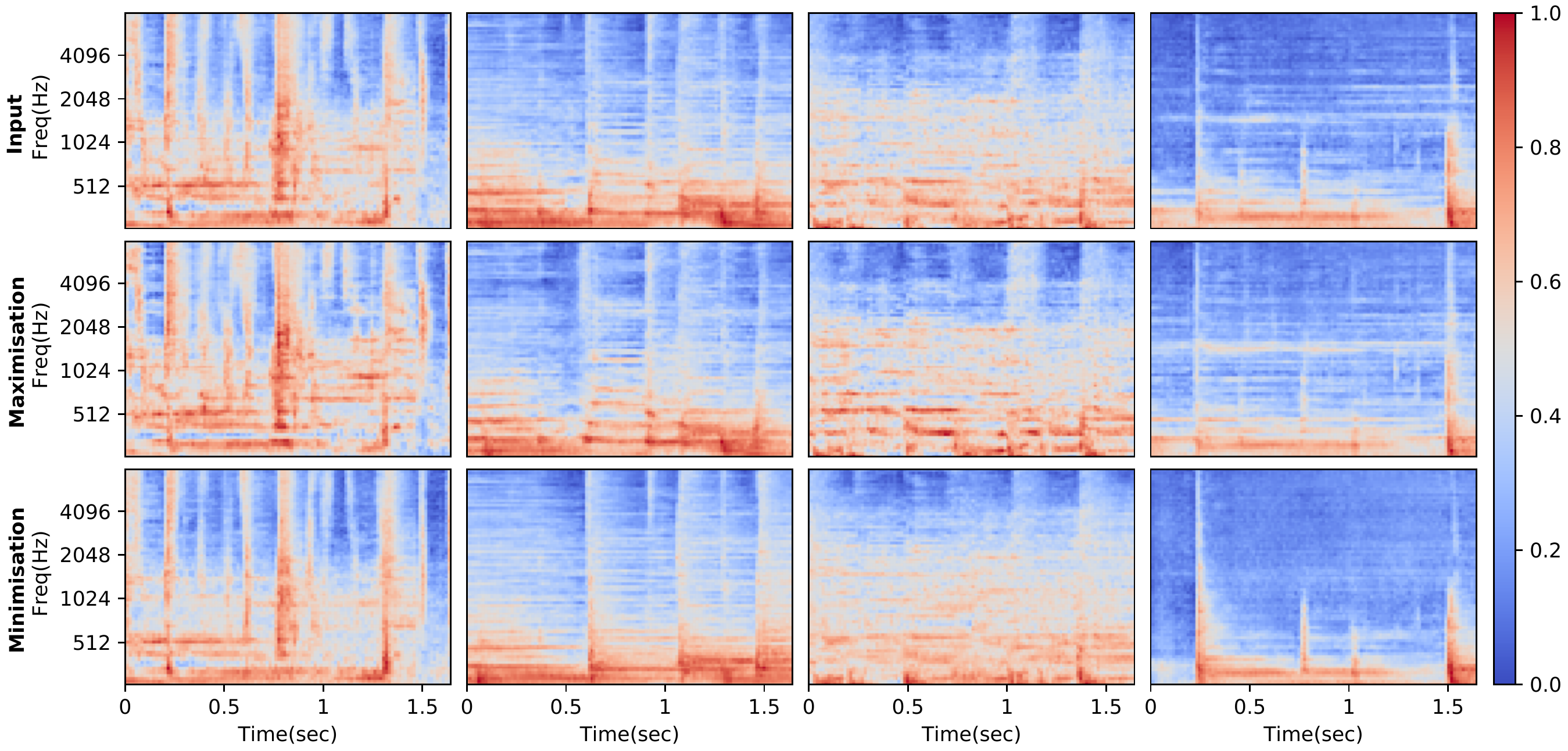}}
\end{center}   
 \caption{Mel spectrogram visualisations illustrating the concepts the neuron in the output layer of the classifier learns, each normalised in scale independently so that red colors show relatively high and blue colors show relatively low spectral energy. The top row represents initial GAN outputs $f_g(\tilde{z}_i)$ for four initial noise vectors $\tilde{z}_1, \tilde{z}_2, \tilde{z}_3, \tilde{z}_4$. The second and third rows represent examples synthesised by maximally and minimally activating the output neuron. Table~\ref{table:classifier} in the Appendix details the activations from the last layer neuron for each visualisation depicted above.}
 \label{fig:classifier}
 \end{figure*}

\subsection{Hyper-parameter optimisation}
\label{sec:results_hyperparam}

We investigate whether our evaluation metric reflects the quality of the results obtained for different hyper-parameter settings. 
Figure~\ref{fig:hyperparam} shows the GAN output for two randomly sampled initial noise vectors $\tilde{z}_1, \tilde{z}_2$ and the corresponding results after maximising the last neuron activation using three different hyper-parameter configurations $C_1$, $C_2$ and $C_3$.
In our best configuration $C_1$, only small changes are made to the initial output, which ensures diverse and realistic outputs due to the random values of $\tilde{z}$ and high likelihood under the prior $p_z$, while still increasing the response effectively.
The median configuration $C_2$ leads to further increased harmonic as well as high-frequency energy, although to a slightly unrealistic extent, and configuration $C_3$ produces extremely sparse outputs with maximum energies one magnitude greater than those found in real spectrograms, and unrealistically high responses of up to $78$, thus is ineffective.
This shows our metric can rank the different hyper-parameter settings with respect to how useful the resulting explanations are for a human observer. Further perceptual studies are left for future work to establish a stronger connection.

\subsection{Qualitative analysis of explanations}
\label{sec:results_classifier}

We use the configuration $C_1$ (Table~\ref{table:hyperparams} in Appendix) to produce positive and negative examples for the last output neuron of our vocal classifier by maximising or minimising its activation, respectively. 
Since the classifier was trained to distinguish vocal from non-vocal audio, this allows us to investigate whether our method can successfully capture the concept of vocal presence using its positive examples, and the concept of pure accompaniment using its negative examples.

Figure~\ref{fig:classifier} shows four pairs of positive and negative examples, each generated using the same noise vector $\tilde{z}$ as initialisation point.
We observe a stronger presence of harmonic content in the positive examples, a lack of energy in the very low frequency band below the human voice range, and few transient sounds such as drum hits (visible as vertical bars), indicating our positive explanations indeed have many characteristics typical to vocal content.
In contrast, the negative examples have stronger transients and more bass frequency content, indicating the successful generation of purely instrumental examples.
Since initial listening tests of the resynthesised audio confirms these observations, this suggests that our GAN-based approach can provide explanations useful for understanding the concepts acquired by a neural network.
Furthermore, Table~\ref{table:classifier} demonstrates that our method effectively optimises the response in all cases.
A more quantitative, large-scale listening test is left for future work.

\section{Conclusions}

In this paper, we presented a GAN-based approach for efficiently generating inputs to a classifier so that its response is maximised, while maintaining realism thanks to its strong prior. Compared to previous approaches, it can be applied more flexibly to new classifiers and to excite different neurons in a classifier.
We validated our method on a pre-trained singing voice classifier, showing it can retrieve the concept of singing voice presence encoded in the last output neuron.
We presented a metric for automatic evaluation of the usefulness of a set of generated explanations, and use it for optimising the hyper-parameters of our approach. We qualitatively showed that our metric favours the subjectively more interpretable settings.
For future work, we plan to conduct listening tests that present generated examples and require the prediction of the model's behaviour for unseen inputs to quantify the interpretability of the explanations.

\subsubsection*{Acknowledgments}

DS is funded by EPSRC grant EP/L01632X/1. EB is supported by RAEng Research Fellowship RF/128 and a Turing Fellowship. This work is supported by EPSRC grant EP/R01891X/1.
\bibliography{iclr2019_conference}
\bibliographystyle{iclr2019_conference}

\newpage
\section*{Appendix}

\begin{table}[h]
\footnotesize
\caption{Performance comparison between the model trained by~\citet{Schluter2015} (``Original") and our replication (``Replication") on the Jamendo test dataset. We can see that both models are very close in their predictive capability.}
\label{table:svdnet_compare}
\begin{center}
\begin{tabular}{ccccccc} 
  \toprule
  Model & Threshold & Precision & Recall & Specificity & F1-score & Classification error\\ 
  \midrule
 Original & 0.47 & 0.901 & 0.926 & 0.912 & 0.913 & 0.082 \\
 Replication & 0.50 & 0.896 &  0.925 & 0.908 & 0.910 & 0.084 \\
  \bottomrule
\end{tabular}
\end{center}
\end{table}

\begin{table}[h]
\footnotesize
\caption{The architecture of SVDNet introduced by \citet{Schluter2015}. Conv, MP and FC refer to the convolutional, max-pooling and fully-connected layers, respectively. Input and output shapes are ordered as: time $\times$ frequency $\times$ number of channels for the Conv layers. }
\label{table:svdnet_arch}
\begin{center}
\begin{tabular}{cccccc} 
 \toprule
 Layer & Input shape & Filter size & Stride & No. of filters/neurons & Output shape\\
 \midrule
 Conv & $ 115 \times 80 \times 1 $ & $3\times 3$ & $1\times 1$ & 64 & $113 \times 78 \times 64$ \\ 
 Conv & $ 113 \times 78 \times 64 $ & $3\times 3$ & $1\times 1$ & 32 &  $111 \times 76 \times 32$ \\
 MP & $ 111 \times 76 \times 32 $ & $3\times 3$ & $3 \times 3$ & - &  $37 \times 25 \times 32 $ \\
 Conv & $ 37 \times 25 \times 32 $ & $3\times 3$ & $1\times 1$ & 128 &  $35 \times 23 \times 128$ \\
 Conv & $ 35 \times 23 \times 128 $ & $3\times 3$ & $1\times 1$& 64 & $33 \times 21 \times 64$ \\
 MP & $ 33 \times 21 \times 64 $ & $3\times 3$ & $3 \times 3$ & - &  $11 \times 7 \times 64$ \\
 FC & $ 11 \times 7 \times 64 $ & - & - & 256 &  $256 \times 1$ \\
 FC & $ 256 $ & - & - & 64 &  $64 \times 1$ \\
 FC & $ 64 $ & - & - &  1 & 1\\ 
 \bottomrule
\end{tabular}
\end{center}
\end{table}

\begin{table}[h]
\footnotesize
\caption{Best, median and worst hyper-parameter configuration found during hyper-parameter search in~\secref{sec:hyperparam}. Columns indicate name, learning rate, GAN prior weight, number of SGD iterations, and the resulting value of our FID-based evaluation metric, respectively.}
\label{table:hyperparams}
\begin{center}
\begin{tabular}{cccccc} 
 \toprule
 Config. name & $l_r$ & $\lambda$ & $N_t$ & FID \\
 \midrule
 C1 & 0.01 & 0.001 & 100 & 1.654 \\
 C2 & 0.01 & 0.01 & 500 & 35.880 \\
 C3 & 0.1 & 0.001 & 1000 & 1557.733 \\
 \bottomrule
\end{tabular}
\end{center}
\end{table}

\begin{table}[h!]
\footnotesize
\caption{Reponse for each Mel spectrogram in Figure~\ref{fig:classifier}. $a_1, a_2, a_3$ and $a_4$ refer to reponse values to the corresponding noise vector.}
\label{table:classifier}
\begin{center}
\begin{tabular}{ccccc} 
 \toprule
 Category &  $a_1$ & $a_2$ & $a_3$ & $a_4$ \\
 \midrule
 $ {f_g}(\tilde{z}_i)$ & -0.5 & -0.94 & 1.55 & -3.47 \\
 Maximisation & 5.83 & 5.52 & 6.1 & 2.96 \\
 Minimisation & -6.23 & -7.21 & -6.45 & -6.36 \\
 \bottomrule
\end{tabular}
\end{center}
\end{table}

\end{document}